\documentclass[conference]{IEEEtran}

\IEEEoverridecommandlockouts
\usepackage{cite}
\usepackage[pdftex]{graphicx}
\usepackage{epstopdf} %included
\graphicspath{{./figures/}{./pictures/}}
\usepackage[cmex10]{amsmath}
\usepackage{array}
\usepackage[small]{caption}
\usepackage{verbatim}
\usepackage{graphicx}% remove the 'demo' option in the real document
\usepackage{balance}
\usepackage{color} % to color the comments

\usepackage{graphicx}% http://ctan.org/pkg/graphicx
\usepackage{stfloats}
\usepackage{amsfonts} 
\usepackage{amssymb} 
\usepackage{url}
\usepackage{booktabs}
\usepackage[utf8]{inputenc}
\usepackage{mathrsfs}
\usepackage{xspace}
\usepackage{xfrac}
\usepackage{caption}
\usepackage{subfig}
\usepackage{multirow}
\usepackage{mathtools, cuted}

\RequirePackage[utf8]{inputenc}
\usepackage{amsmath}
\usepackage{booktabs}
\newcommand{\R}{{\mathbb R}} 
\renewcommand{\S}{{\mathbb S}} 
\newcommand{\Z}{{\mathbb Z}}

\renewcommand{\H}{{\mathcal{H}}}

\newcommand{\0}{\boldsymbol{0}}

\newcommand{\vv}{\boldsymbol{v}}
\newcommand{\dT}{\Delta T}

\newcommand{\G}{\text{G}}
\newcommand{\F}{\mathcal{F}}
\newcommand{\g}{\mathfrak{g}}
\newcommand{\SE}{\text{SE}}
\newcommand{\se}{\mathfrak{se}}
\newcommand{\SO}{\text{SO}}

\newcommand{\x}{\boldsymbol{x}}
\newcommand{\I}{\text{I}}

\renewcommand{\t}{\boldsymbol{t}}

\DeclareMathOperator{\atan2}{atan2}
\DeclareMathOperator{\ad}{ad}
\DeclareMathOperator{\Ad}{Ad}

\begin{document}

% *** PAPER TITLE ***
%
\title{ 
Moving object tracking employing rigid body motion on matrix Lie groups\vspace{-0mm}
}
\author{\IEEEauthorblockN{Josip Ćesić, Ivan Marković, Ivan Petrović}
\IEEEauthorblockA{University of Zagreb, Faculty of Electrical Engineering and Computing, Unska 3, 10000 Zagreb, Croatia\\
Email: \texttt{josip.cesic@fer.hr, ivan.markovic@fer.hr, ivan.petrovic@fer.hr}}
}

\maketitle

\begin{abstract}
In this paper we propose a novel method for estimating rigid body motion by modeling the object state directly in the
space of the rigid body motion group $\SE(2)$.
It has been recently observed that a noisy manoeuvring object in $\SE(2)$ exhibits \textit{banana-shaped} probability
density contours in its pose.
For this reason, we propose and investigate two state space models for moving object tracking: (i) a direct product
$\SE(2)\times\R^3$ and (ii) a direct product of the two rigid body motion groups $\SE(2)\times\SE(2)$.
The first term within these two state space constructions describes the current pose of the rigid body, while the second
one employs its second order dynamics, i.e., the velocities.
By this, we gain the flexibility of tracking omnidirectional motion in the vein of a constant velocity model, but also
accounting for the dynamics in the rotation component.
Since the $\SE(2)$ group is a matrix Lie group, we solve this problem by using the extended Kalman filter on matrix Lie
groups and provide a detailed derivation of the proposed filters.
We analyze the performance of the filters on a large number of synthetic trajectories and compare them with
(i) the extended Kalman filter based constant velocity and turn rate model and (ii) the linear Kalman filter based
constant velocity model.
The results show that the proposed filters outperform the other two filters on a wide spectrum of types of motion.
%\im{Dodati zaključak eksperimenata, reći kako predloženi filtar ostvaruje najbolje rezultate.}
%\im{In order to tackle real-world tracking problems with clutter, we also propose the framework for employing the
%probabilistic data association (PDA) on matrix Lie Groups.
%We illustrate the application thereof on a mobile robot tracking problem with different levels of clutter rate.}
\end{abstract}

%\begin{keywords}
%Membrain, Model Predictive Control, Fault tolerant, Piece-wise affine, Phase-change Material
%\end{keywords}

%\IEEEpeerreviewmaketitle

\section{Introduction}
\label{sec:intro}

A wide area of robotics research has extensively focused on the practical approaches of using different types of
manifolds.
Besides performance, filters operating on manifolds can provide other advantages as they avoid singularities when
representing state spaces with either redundant degrees of freedom or constraint issues
\cite{Barfoot2014,Hertzberg2013}.
% which mostly results from the fact that the members of different manifolds, as well as $\SE(2)$ members, are not vectors, but rather noncommutative matrices \cite{Barfoot2014}.
Among the manifolds, the homogeneous transformation matrices, also referred to as the rigid body motion group
$\SE(n)$, hold a special repute.
They have been used in a variety of applications, and have risen to popularity firstly through manipulator robotics
\cite{Murray1994,Park1995} and later through vision applications \cite{Davison2003,Lui2012}.
Even though the state description using the rigid body motion group, for both the $2$D and $3$D case, has been a
well known representation, techniques for associating the uncertainty came into focus later \cite{Lee1992}.
So far, the rigid body motion group with associated uncertainty has been used in several robotics applications such
as SLAM \cite{Silveira2008}, motion control \cite{Park2010a}, shape estimation \cite{Srivatsan2014}, pose estimation \cite{Long2012} and pose registration \cite{Agrawal2006}.

\begin{figure}[t]
\centering
\includegraphics[width=1\columnwidth]{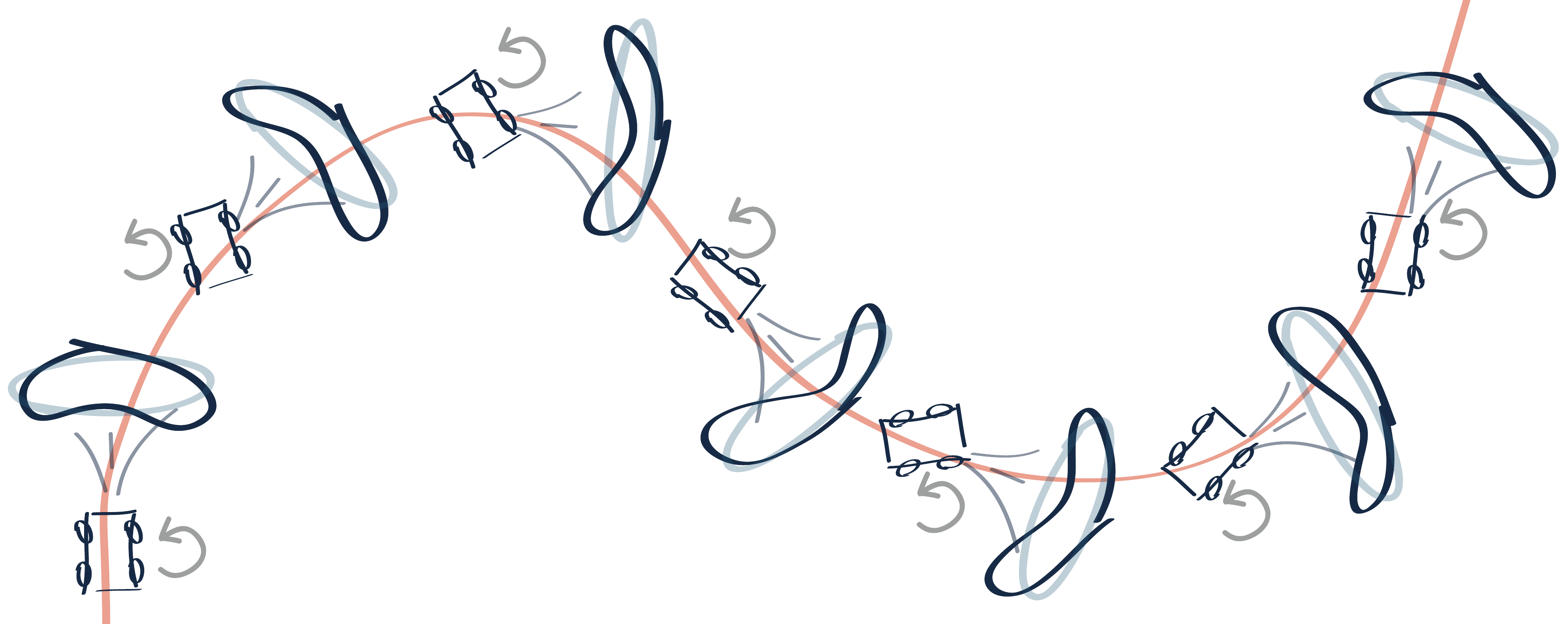}
\caption{An illustration of an omnidirectional mobile robot manoeuvring in both translational and rotational components.
The banana shaped uncertainty contours, representing the positional uncertainty in the next step,
are formed by modeling the uncertainty on the $\SE(2)$ group (blue), 
while the elliptical shaped contours appear modeling the uncertainty in $\R^2$ (gray).}
\vspace{-5mm}
\label{fig:state_space_icra}
\end{figure}

Among them, pose estimation represents one of the central problems in robotics. 
Recently in \cite{Long2012} the authors discussed the advantages of employing uncertainties on $\SE(2)$ (therein called
the exponential coordinates) with respect to Euclidean spaces and have provided the means for working in the exponential
coordinates rather than representing the robot's position with Gaussians in Cartesian coordinates. 
This stems from the fact that the uncertain robot motion, and consequently its pose, usually exhibit
\textit{banana-shaped} probability density contours rather than the elliptical ones \cite{Thrun2006}, as illustrated
in Fig.~\ref{fig:state_space_icra}.
The classical Kalman filter is designed to operate in the Cartesian space and as such does not provide a framework for filtering directly on the $\SE(2)$ group.
%Thus, the conventional approaches are incapable of exploiting any-order dynamics directly in the
%exponential coordinates.
Recently, some works have addressed the uncertainty on the $\SE(2)$ group proposing new distributions \cite{Kurz2014a,Gilitschenski2014}.
However, these interesting approaches do not yet provide a closed-form Bayesian recursion framework (involving both the
prediction and update) that can include higher order motion and non-linear models.

An extended Kalman filter on matrix Lie groups (LG-EKF) has been recently proposed in \cite{Bourmaud2014}. 
It provides an estimation framework for filtering directly on matrix Lie groups, of which the $\SE(2)$ group is a
member. 
%Furthermore, the LG-EKF enables to compose the state space by combining different types of Lie groups using different products between spaces (e.g. direct product, semi-direct product, twisted product, etc.). 
%Still, the general choice of the product is still an open question and as such is left on the choice of practitioner \cite{Bourmaud2014}. 
In accordance with the needs of moving object state estimation problems, higher order motion often
needs to be exploited, as in the vein of the constant velocity (CV) or acceleration motion models
\cite{Bar-Shalom2002}, but in the space such as the rigid body motion group $\SE(2)$.
In the present paper we propose a method for moving object tracking employing its second order motion
directly on the $\SE(2)$ group based on the discrete LG-EKF. 
For this purpose, we model the state space either as a direct product of (i) a rigid body motion group and a Euclidean vector or (ii) two rigid body motion groups, i.e.,
\begin{equation}
\text{(i) } \SE(2)\times\R^3 \text{ \ or \ }\text{(ii) } \SE(2) \times \SE(2) = \SE(2)^2.
\end{equation}
In both cases the first term tracks the pose of the object, while the second one handles the velocities.
%Furthermore, we present the means for utilization of manoeuvring constraints for tracking of a moving object in the same $\SE(2) \times \SE(2)$ state space.
In the end, we conduct experimental validation of the proposed filters on synthetic data and compare their performance
with the CV and constant turn rate and velocity (CTRV) motion models \cite{Schubert2012} used within the classical
extended Kalman filter (EKF) framework.
%\im{The second contribution of this paper is a framework for employing the
%probabilistic data association (PDA) method \cite{Bar-Shalom1975} on matrix Lie Groups, in order to tackle real-world tracking problems with clutter.
%The importance of the presented derivation is manifold since this serves as a basis for many data association based approaches for single \cite{Musicki1994} and multi-target tracking \cite{Fortmann1983,Musicki2004} applications.
%We illustrate the behaviour of the proposed filter in a simple real-world experiment, applying it on an example of mobile robot tracking with a 3D laser range sensor.}

%\begin{comment}
The rest of the paper is organized as follows. 
Section \ref{sec:mot} gives an insight into the motivation behind the present paper.
Section \ref{sec:pr} provides the preliminaries including the basic definitions and operators for working with matrix
Lie groups, with emphasis on the special euclidean group $\SE(2)$. The method for exploiting higher order motion is
presented in Section \ref{sec:SE2} and the proposed estimation strategies are investigated on a synthetic dataset and
compared with two Kalman filter based methods. Finally, concluding remarks are drawn in Section \ref{sec:con}.
%\end{comment}

\section{Motivation}
\label{sec:mot}
The choice of the state space and the approach to the motion modelling present a significant focus of this paper. 
The physical interpretation behind associating the uncertainty with the $\SE(2)$ group has been analyzed in
\cite{Long2012}.
Therein, the authors particularly study the shape of the uncertainty by considering differential drive mobile robot
motion.
The authors conclude that the $\SE(2)$ approach provides significant flexibility in describing the position uncertainty,
enabling one to analytically work with banana-shaped uncertainty contours.
In this work, given the previous moving object tracking discussion, we aim to track omnidirectional motion in order to
achieve high flexibility in motion modeling.
This is motivated by considering tracking in unknown dynamic environments comprising of multiple unknown moving objects.
For example, a mobile robot building a map of an unknown environment consisting of humans and other robots with various
kinematics, or a busy intersection with mixed traffic involving cars, trams, motorcycles, bicycles and pedestrians.

By searching for the flexibility to control the velocities in both $x$ and $y$ direction, as well as the rotational
velocity, one comes to formulation of the state space as $\SE(2) \times \R^3$.
In this case, the $\SE(2)$ term tracks the pose of a rigid body object supporting the forming of
banana-shaped uncertainty contours, while the $\R^3$ term describes velocities along the three axis in a classical manner
forming elliptical-only contours.
Examples of omnidirectional mechanical robot platforms implementations which can be described by this state space
construction are the \textit{Palm Pilot Robot}, \textit{Uranus}, and \textit{Killough} \cite{Siegwart2004}, which
are based on the Swedish $45^{\circ}$/$90^{\circ}$ wheels.

However, if we consider a robot construction that has additional flexibility of controling the steering angle of one or
more wheels, it turns out that by sampling such kinematic models the uncertainty in the space of velocities also has
banana-shaped contours.
Given that, we further propose to model the state space as $\SE(2)^2$ group where now the second term exploits the
second order motion (velocities), and supports the flexibility of forming the banana-shaped uncertainty contours
in the velocity space.
Examples of mechanical omnidirectional robot platforms capable of such motion are the \textit{Nomad XR4000}
and \textit{Hyperion} \cite{Siegwart2004}.
Detailed physical and kinematic interpretations of these models are, however, out of the scope of this paper
and are a subject for future work.

%In the sequel we provide the preliminaries needed for 
%derivation of the tracking approach
%based on the LG-EKF filter and the probabilistic data association concept.

\section{Preliminaries}
\label{sec:pr}
\subsection{Lie groups and Lie algebra}
\label{sec:prelim}

In this section, we provide notations and properties for matrix Lie groups and the associated Lie algebras which will be
used for the filter including the $\SE(2)$ group in the state space.
\begin{comment}
Here we focus on matrix Lie groups since following Ado's theorem any finite dimensional Lie group is homomorphic to a matrix Lie group \cite{Ado1947}.
Furthermore, often several representations of the same physical manifold are possible (e.g. $\R/2\pi\Z$, $\SO(2)$ and $\S^1$ describe $2$D rotations\footnote{The directions in $\R^{n+1}$ are represented as unit vectors in $\S^n$, forming thus the unit sphere $\S^n = \{ x\in \R^{n+1} \ | \ ||x||=1 \}$.}), among which some belong to family of matrix Lie groups.
\end{comment}
For a more formal introduction of the used concepts, the interested reader is directed to \cite{Chirikjian2012b}, where the
author provides a rigorous treatment of representing and propagating uncertainty on matrix Lie groups.

The $\SE(2)$, specifically, is a matrix Lie group.
A Lie group is a group which has the structure of a smooth manifold, i.e., it is sufficiently often differentiable
\cite{Hertzberg2013}, such that group composition and inversion are smooth operations.
Furthermore, for a matrix Lie group $\G$ these operations are simply matrix multiplication and inversion, with the identity matrix $\I^{n \times n}$ being the identity element \cite{Chirikjian2012b}.
An interesting property of Lie groups, basically curved objects, is that they can be almost completely captured by a
flat object, such as the tangential space; and this leads us to an another important concept---the Lie algebra $\g$
associated to a Lie group $\G$.

Lie algebra $\g$ is an open neighborhood of $\0^{n \times n}$ in the tangent space of $\G$ at the identity $\I^{n \times n}$.
The matrix exponential $\exp_{\G}$ and matrix logarithm $\log_{\G}$ establish a local diffeomorphism between these two
worlds, i.e., Lie groups and Lie algebras
\begin{align}
\exp_{\G} : \g \rightarrow \G \text{ \ and \ }
\log_{\G} : \G \rightarrow \g.
\end{align}
The Lie algebra $\g$ associated to a $p$-dimensional matrix Lie group $\G \subset \R^{n \times n}$ is a $p$-dimensional vector space defined by a basis consisting of $p$ real matrices $E_i$, $i=1,..,p$ \cite{Park2010a}.
A linear isomorphism between $\g$ and $\R^p$ is given by
\begin{align}
[\cdot]^{\vee}_{\G}  : \g \rightarrow \R^p \text{ \ and \ }
[\cdot]^{\wedge}_{\G}  : \R^p \rightarrow \g.
\end{align}
Lie groups are not necessarily commutative and require the use two operators to capture this property and thus, enable
the \emph{adjoint representation} of (i) $\G$ on $\R^p$ denoted as $\Ad_{\G}$ and (ii) $\R^p$ on $\R^p$ denoted as $\ad_{\G}$
\cite{Chirikjian2012b}.
All the discussed operators in the present section are presented later in the paper for the proposed state space
constructions.

\begin{comment}
\begin{align}
\Ad_{\G} & : \Ad_{\G}(X) a  =
\left[ X [a]^{\wedge}_{\G} X^{-1} \right]^{\vee}_{\G}\\
\ad_{\G} & : \ad_{\G}(a) b  =
\left[ [a]^{\wedge}_{\G} [b]^{\wedge}_{\G} - [b]^{\wedge}_{\G} [a]^{\wedge}_{\G}\right]^{\vee}_{\G} \,,
\end{align}
where $a,b\in \R^p$.
\end{comment}

\begin{comment}
One of the main tools that is used to manipulate the uncertainties on matrix Lie groups is the Baker-Campbell-Hausdorff (BCH) formula used to compound two matrix exponentials which expresses the group product directly in $\R^p$
\begin{equation}
\begin{split}
& \left[ \log_{\G} \left[ \exp_{\G} \left( [g_1]^{\wedge}_{\G} \right) 
\exp_{\G} \left( [g_2]^{\wedge}_{\G} \right) \right] \right]^{\vee}_{\G}=\\
& \qquad \qquad = g_1+g_2+ [g_1,g_2]+\mathcal{O}(|g_1,g_2|^3) \,.
\end{split}
\end{equation}
where $[\cdot,\cdot]$ is so called the Lie bracket and is given with $[g_1,g_2]:=g_1 g_2 - g_2 g_1$.
Another useful relation used further in filter derivation is 
\begin{equation}
\begin{split}
&\left[
\log_{\G}
\left[
\exp_{\G} \left( [-g_1]^{\wedge}_{\G} \right)
\exp_{\G} \left( [g_1+g_2]^{\wedge}_{\G} \right)
\right]
\right]^{\vee}_{\G} =\\
& \qquad \qquad = g_1+\Phi_{\G}(g_1)g_2+\mathcal{O}(|g_2|^2) \,,
\end{split}
\end{equation}
where
\begin{align}
\Phi_{\G}(g)= \sum_{m=0}^{\infty} \dfrac{(-1)^m}{(m+1)!} \ad_{\G}(g)^m  \,.
\end{align}
is called the right Jacobian of $\G$ \cite{Barfoot2014}.
\end{comment}

\subsection{Concentrated Gaussian Distribution}

Another important concept in the LG-EKF framework is that of the concentrated Gaussian distribution (CGD).
In order to define the CGD on matrix Lie groups, the considered group needs to be a connected unimodular matrix Lie
group \cite{Wang2008}, which is the case for the majority of martix Lie groups used in robotics. 

Let the probability density function (pdf) of $X$, a state on a $p$-dimensional matrix Lie group $\G$, be defined as
\cite{Wolfe2011}
\begin{align}\label{eq:cgd}
  p(X) = \beta \exp\left({-\dfrac{1}{2} [\log_{\G}(X)]^{\vee^T}_{\G}} P^{-1} [\log_{\G}(X)]^{\vee}_{\G}\right) \,,
\end{align}
where $\beta$ is a normalizing constant chosen such that \eqref{eq:cgd} integrates
to unity.
In general $\beta \neq (2 \pi)^{-p/2} | P |^{-1/2}$ with $|\cdot|$ being the matrix determinant and $P$ a positive
definite matrix.

Furthermore, let $\epsilon$ be defined as $\epsilon \triangleq [\log_{\G}(X)]^{\vee}_{\G}$.
If we now assume that the entire mass of probability is contained inside $\G$,
%, i.e. $\int_{\R^{n \times n} \backslash \G}p(X)=0$
then $\epsilon$ can be described by $\epsilon \sim \mathcal{N}_{\R^p}(\0^{p \times 1},P)$.
This represents the CGD on $\G$ around the identity \cite{Bourmaud2014}.
Furthermore, it is a unique parametrization space where the bijection between $\exp_{\G}$ and $\log_{\G}$ exists. 
Now, the pdf of $X$ can be `translated' over the $\G$ by using the left action of the matrix Lie group
\begin{align}
X=\mu \exp_{\G}  \left( [\epsilon]^{\wedge}_{\G} \right) \ \text{, with} \ \
X \sim \mathcal{G}(\mu,P)\,,
\end{align}
where $\mathcal{G}$ denotes the concentrated Gaussian distribution \cite{Wolfe2011,Bourmaud2014} with the mean $\mu$ and
the covariance matrix $P$.
In other words, the mean $\mu$ of the state $X$ resides on the $p$-dimensional matrix Lie group $\G$, while the
associated uncertainty is defined in the space of the Lie algebra $\g$, i.e., by the linear isomorphism the Euclidean
vector space $\R^p$.
By this, we have introduced the distribution forming the base for the LG-EKF.

\subsection{The $\SE(2)$ group}

The motion group $\SE(2)$ describes the rigid body motion in $2$D and is formed as a semi-direct product of the plane $\R^2$ and the special orthogonal group $\SO(2)$ corresponding to translational and rotational parts, respectively. It is defined as
\begin{equation}
\SE(2) = \left\{ \begin{pmatrix}
R & \t\\
\0^{1\times 1} &  1
\end{pmatrix}
\in \R^{3 \times 3} \, | \, 
\{ R , \t \} \in \SO(2) \times \R^2
\right\} \,.
\end{equation}
Now, we continue with providing the basic ingredients for handling $\SE(2)$, giving the relations for operators from \ref{sec:prelim}, needed for manipulation between the triplet ({Lie group} $\G$, {Lie algebra} $\g$, {Euclidean space} $\R^p$).

For the Euclidean spaced vector 
$\x = \begin{bmatrix}
x \ y \ \theta
\end{bmatrix}^T$,
the most often associated element of the Lie algebra $\se(2)$ is given as
\begin{align}
[\x]^{\wedge}_{\SE(2)} = 
\begin{bmatrix}
0 & -\theta & x\\
\theta & 0 &  y\\
0 & 0 & 0
\end{bmatrix} \in \se(2) \,.
\end{align}
Correspondingly, its inverse $[\cdot]^{\vee}_{\SE(2)}$ is trivial.

The exponential map for the $\SE(2)$ group is given as
\begin{align}
&\exp_{\SE(2)}([\x]^{\wedge}_{\G}) = 
\begin{bmatrix}
\cos\theta & -\sin\theta & t_x\\
\sin\theta & \cos\theta & t_y\\
0 & 0 & 1
\end{bmatrix} \in \SE(2) \\
& \qquad t_x = \frac{1}{\theta} \left[ x\sin\theta + y(-1+\cos\theta) \right] \\
& \qquad t_y = \frac{1}{\theta} \left[ x(1-\cos\theta) + y\sin\theta) \right] \,.
\end{align}
%\begin{comment}
For $T=\{R,\t\} \in \SE(2)$, the logarithmic map is
\begin{align}
& \log_{\SE(2)}(T)  = 
\begin{bmatrix}
\vv \\ \theta
\end{bmatrix}
^{\wedge}_{\SE(2)} \in \se(2)\\
& \qquad \theta = \log_{\SO(2)}(R) = \atan2(R_{21},R_{11})\\
& \qquad \vv = \dfrac{\theta}{2(1-\cos\theta)}
\begin{bmatrix}
\sin\theta & 1-\cos\theta \\
\cos\theta-1 & \sin\theta
\end{bmatrix} \t \,.
\end{align}
%\end{comment}

The Adjoint operator $\Ad_{\G}$ used for representing $T \in \SE(2)$ on $\R^3$ is given as
\begin{align}
\Ad_{\SE(2)}(T) = 
\begin{bmatrix}
R & J \t\\
\0^{1 \times 2} & 1
\end{bmatrix} \ \text{with} \
J =
\begin{bmatrix}
0 & 1\\
-1 & 0
\end{bmatrix} \,.
\end{align}
The adjoint operator $\ad_{\G}$ for representing $\x \in \R^3$ on $\R^3$ is given by
\begin{align}
\ad_{\SE(2)}(\x) = 
\begin{bmatrix}
-\theta J & J \vv \\
\0^{1 \times 2} & 1
\end{bmatrix} \,,
\end{align}
where $ \vv =[x \ y]^T \in \R^2$.

\section{Rigid body motion tracking}
\label{sec:SE2}
%\subsection{Basics on rigid body motion group $\SE(2)$}

\subsection{EKF on matrix Lie groups}
\label{sec:filter}

For the general filtering approach on matrix Lie groups, the system is assumed to be modeled as satisfying the following
equation
\cite{Bourmaud2013a}
\begin{equation}
\begin{split}
X_{k+1}
 = f(X_{k},n_{k})
 = X_{k} \, \exp_{\G} \left( [\hat{\Omega}_k + n_{k}]^{\wedge}_{\G} \right) \,,
\end{split}
\label{eq:model}
\end{equation}
where $X_k \in \G$ is the state of the system at time $k$, $\G$ is a $p$-dimensional Lie group, $n_k \sim
\mathcal{N}_{\R^p}(\0^{p \times 1},Q_k)$ is white Gaussian noise and $\hat{\Omega}_k=\Omega(X_{k}):\G \rightarrow
\R^p$ is a non-linear $\mathcal{C}^2$ function.

The prediction step of the LG-EKF, based on the motion model \eqref{eq:model}, is governed by the following formulae
\begin{align}
\label{eq:pred_mu}
\mu_{k+1|k} & = \mu_{k} \exp_{\G}  \left( [\hat{\Omega}_k]^{\wedge}_{\G} \right)\\
\label{eq:pred_cov}
P_{k+1|k} & = \F_{k} P_k \F_{k}^T + \Phi_{\G}(\hat{\Omega}_{k}) Q_k \Phi_{\G}(\hat{\Omega}_{k})^T \,,
\end{align}
where $\mu_{k+1|k} \in \G$ and $P_{k+1|k} \in \R^{p \times p}$ are predicted mean value and the covariance matrix, respectively, hence the state remains $\mathcal{G}$--distributed $X_{k+1|k} \sim \mathcal{G}(\mu_{k+1|k},P_{k+1|k})$.
The operator $\F_{k}$, a matrix Lie group equivalent to the Jacobian of $f(X_k,n_k)$, and $\Phi_{\G}$ are given as follows
\begin{align}
\F_k & = \text{Ad}_{\G} \left( \exp_{\G} \left([-\hat{\Omega}_k]^{\wedge}_{\G} \right) \right) + \Phi_{\G}(\hat{\Omega}_{k}) \mathscr{C}_{k} \\
\Phi_{\G}(\vv) & = \sum_{m=0}^{\infty} \dfrac{(-1)^m}{(m+1)!} \ad_{\G}(\vv)^m\,, \ \vv \in \R^p  \\
\label{eq:C}
\mathscr{C}_{k} & = \dfrac{\partial}{\partial \epsilon} \Omega \left( \mu_{k} \exp_{\G}  \left( [\epsilon]^{\wedge}_{\G} \right) \right)_{|\epsilon=0}  \,.
\end{align}

\begin{figure*}[!t]
\centering
%  \tikzsetnextfilename{bananas}
%  \input{figures/bananas.tikz.tex}
\includegraphics[width=1\textwidth]{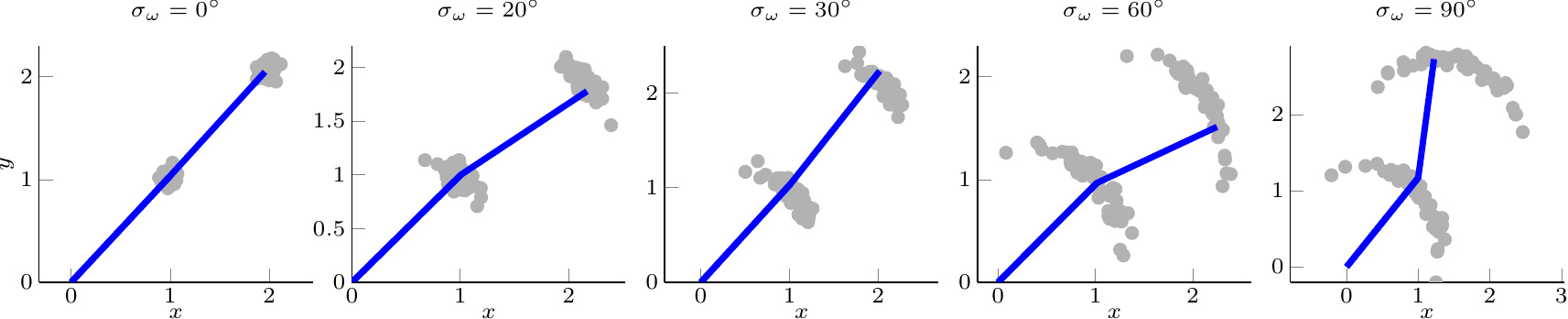}
\caption{
Each of the subfigures represents an example of two compounding transformations for different levels of rotational uncertainty (given in blue). The grey circles represent $50$ sampled uncertain transformations by employing both translational and rotational uncertainties.
This particular situation appears when a robot moves from the current position to the next position associated with the next discrete moment in time, with standard deviation of the rotation $\sigma_{\omega}$.
\vspace{-6mm}
}
\label{fig:bananas}
\end{figure*}

The discrete measurement model on the matrix Lie group is modelled as
\begin{align}
Z_{k+1} = h(X_{k+1}) \, \exp_{\G'} \left( [m_{k+1}]^{\wedge}_{\G'} \right) \,,
\label{eq:meas}
\end{align}
where $Z_{k+1} \in \G'$, $h:\G \rightarrow \G'$ is a $\mathcal{C}^1$ function and $m_{k+1} \sim \mathcal{N}_{\R^q}(\0^{q
\times 1},R_{k+1})$ is white Gaussian noise.

The update step of the filter, based on the measurement model \eqref{eq:meas}, strongly resembles the standard EKF update procedure, relying on the Kalman gain $K_{k+1}$ and innovation vector $\nu_{k+1}$ calculated as follows
\begin{align}
\label{eq:K}
  K_{k+1} &= P_{k+1|k} \H_{k+1}^T \left( \H_{k+1} P_{k+1|k} \H_{k+1}^T + R_{k+1} \right)^{-1}\nonumber\\
  \nu_{k+1} & = K_{k+1} \left( \left[ \log_{\G'} \left( h(\mu_{k+1|k})^{-1}Z_{k+1} \right) \right]^{\vee}_{\G'} \right)
  \,.
\end{align}
The matrix $\H_k$ can be seen as the measurement matrix of the system, i.e., a matrix Lie group equivalent to the
Jacobian of $h(X_k)$, and is given as
\begin{equation}\label{eq:calH}
  \begin{split}
    \H_{k+1} = \dfrac{\partial}{\partial \epsilon}
    &\left[
    \log_{\G'} 
    \left(
    h(\mu_{k+1|k})^{-1}\right.\right.\\
    &\left.\left. h\left( \mu_{k+1|k} \exp_{\G}  \left( [\epsilon]^{\wedge}_{\G} \right)
    \right)
    \right)
    \right]^{\vee}_{\G \ | \epsilon=0} \,.
    \end{split}
\end{equation}
Finally, having defined all the constituent elements, the update step is calculated via
\begin{align}
\label{eq:up_mu}
\mu_{k+1} & = \mu_{k+1|k} \exp_{\G}  \left( [\nu_{k+1}]^{\wedge}_{\G} \right) \\
\label{eq:up_P}
P_{k+1} & = \Phi_{\G}(\nu_{k+1}) 
\left(
\I^{p \times p} - K_{k+1} \H_{k+1}\right)
 P_{k+1|k}
\Phi_{\G}(\nu_{k+1})^T \,.
\end{align}
As in the case of the prediction step, the state $X_{k+1} \sim \mathcal{G}(\mu_{k+1},P_{k+1})$ remains
$\mathcal{G}$--distributed after the correction as well.
For a more formal derivation of the LG-EKF, the interested reader is referred to \cite{Bourmaud2014}.

Since the employment of the $\SE(2) \times \R^3$ follows the similar, but slightly simpler derivation,
in the sequel we derive the LG-EKF filter for estimation on the state space modelled as $\SE(2)^2$. 
This approach is in our case applied, but not limited, to the problem of moving object tracking.

\subsection{LG-EKF on $\SE(2)^2$}
\label{sec:SE2SE2}

As mentioned previously, we model the state $X$ to evolve on the matrix Lie group $\G=\SE(2)^2$ which is symbolically represented by
\begin{align}
X = 
\begin{bmatrix}
\begin{bmatrix}
R_{\theta} & \t\\
\0^{1\times 2} &  1
\end{bmatrix}
&  \\
& 
\begin{bmatrix}
R_{\omega} & \t_v\\
\0^{1\times 2} &  1
\end{bmatrix}
\end{bmatrix}
=
\begin{pmatrix}
T_s\\[1mm]
T_d
\end{pmatrix}_{\G}  \,,
\end{align}
where $T_s$ is the stationary component and $T_d$ brings the second order dynamics. 
%If $\G$ was set as $\G=\SE(2) \times \R^3$, the symbolical representation would as well be constructed block diagonally, but the second element would correspond to the matrix Lie group representation of the Euclidean space.
%This representation is discussed into more details within the paragraph \ref{par:update}.
Note that the matrix Lie group composition and inversion are simple matrix multiplication and inversion, hence the previous symbolic representation can be used for all the calculations dealing with operations on $\G$. 

The Lie algebra associated to the Lie group $\G$ is denoted as $\g=\se(2)^2$, thereby for
$\x = \begin{bmatrix}
\x_p \ \x_d
\end{bmatrix}^T \in \R^6$, where
$\x_p = \begin{bmatrix} x \ y \ \theta \end{bmatrix}^T$ and 
$\x_d = \begin{bmatrix} v_x \ v_y \ \omega\end{bmatrix}^T$,
the following holds
\begin{align}
[\x]^{\wedge}_{\G} = 
\begin{bmatrix}
\left[ \x_p \right]^{\wedge}_{\SE(2)} & \\
 & \left[ \x_d \right]^{\wedge}_{\SE(2)}
\end{bmatrix} =
\begin{pmatrix}
\left[ \x_p \right]^{\wedge}_{\SE(2)}\\[2mm]
\left[ \x_d \right]^{\wedge}_{\SE(2)}
\end{pmatrix}_{\g}.
\end{align}
The exponential map for such defined $\G$ is
\begin{align}
\exp_{\G}([\x]^{\wedge}_{\G}) = 
\begin{pmatrix}
\exp_{\SE(2)} \left(\left[\x_p \right]^{\wedge}_{\SE(2)}\right)\\[2mm]
\exp_{\SE(2)} \left(\left[\x_d \right]^{\wedge}_{\SE(2)}\right)
\end{pmatrix}_{\G} \,.
\end{align}
Now, we have all the necessary ingredients for deriving the terms to be used within the LG-EKF.
Several examples of the uncertain transformations following the $\SE(2)^2$ motion model are shown in Fig.~\ref{fig:bananas} (the $\SE(2)\times \R^3$ model would exhibit similar behaviour). 

\subsubsection{Prediction}
We propose to model the motion \eqref{eq:model} of the system by
\begin{align}
\Omega(X_{k}) & = 
\begin{bmatrix}
T v_{x_k} \
T v_{y_k} \
T \omega_k \
0 \
0 \
0 \
\end{bmatrix}^T \in \R^6 \,, \\ %\text{and} \
n_{k} & = 
\begin{bmatrix}
\frac{T^2}{2} n_{x_k} \
\frac{T^2}{2} n_{y_k} \
\frac{T^2}{2} n_{\omega_k} \
T n_{x_k} \
T n_{y_k} \
T n_{\omega_k} \
\end{bmatrix}^T \in \R^6
\,. \nonumber
\label{eq:motionG}
\end{align}
With such a defined motion model, the system is corrupted with white noise over three separated components, i.e.,
$n_{x}$ the noise in the local $x$ direction, $n_{y}$ the noise in the local $y$ direction and $n_{w}$ as the noise in
the rotational component. 
Given that, the intensity of the noise components acts as acceleration
over the associated axes in the system.
If the system state at the discrete time step $k$ is described with $X_{k} \sim
\mathcal{G}(\mu_{k},P_{k})$, the mean value and the covariance can be propagated using \eqref{eq:pred_mu} and
\eqref{eq:pred_cov}.

The covariance propagation is more challenging, since it requires the calculation of \eqref{eq:C}. 
For the Lie algebraic error 
$ \epsilon \triangleq
\begin{bmatrix}
\epsilon_x \ \epsilon_y \ \epsilon_{\theta} \ \epsilon_{v_x} \ \epsilon_{v_y} \ \epsilon_{\omega}
\end{bmatrix} $,
we need to set the following
\begin{equation}
\begin{split}
&\Omega \left( \mu_{k} \exp_{\G}  \left( [\epsilon]^{\wedge}_{\G} \right) \right) \\
& \hspace{0.1cm} = \begin{bmatrix}
\dT v_{x_k} + 
\dT \cos \omega_k \,v_1
%\dfrac{\epsilon_{v_x} \sin \epsilon_{\omega} + \epsilon_{v_y} (\cos \epsilon_{\omega} - 1)}{\epsilon_{\omega}}
- \dT \sin \omega_k \,v_2
%\dfrac{\epsilon_{v_x} (1 - \cos \epsilon_{\omega}) + \epsilon_{v_y} \sin \epsilon_{\omega}}{\epsilon_{\omega}} 
\\
\dT v_{y_k} + 
\dT \sin \omega_k \,v_1
%\dfrac{\epsilon_{v_x} \sin \epsilon_{\omega} + \epsilon_{v_y} (\cos \epsilon_{\omega} - 1)}{\epsilon_{\omega}}
 + \dT \cos \omega_k \,v_2
%\dfrac{\epsilon_{v_x} (1 - \cos \epsilon_{\omega}) + \epsilon_{v_y} \cos \epsilon_{\omega}}{\epsilon_{\omega}} 
\\
\dT \omega_k + \dT \epsilon_{\omega} \\
\0^{3 \times 1}
\end{bmatrix}  \,.
\label{eq:motionG}
\end{split}
\end{equation}
where
\begin{equation}
\begin{split}
v_1 &= \left[\epsilon_{v_x} \sin \epsilon_{\omega} + \epsilon_{v_y} (\cos \epsilon_{\omega}-1)\right]\epsilon_{\omega}^{-1}\\
v_2 &= \left[\epsilon_{v_x} (1 - \cos \epsilon_{\omega}) + \epsilon_{v_y} \sin \epsilon_{\omega}\right]\epsilon_{\omega}^{-1}.
\end{split}
\label{eq:vs}
\end{equation}
\begin{figure*}[!t]
\centering
\includegraphics[width=1\textwidth]{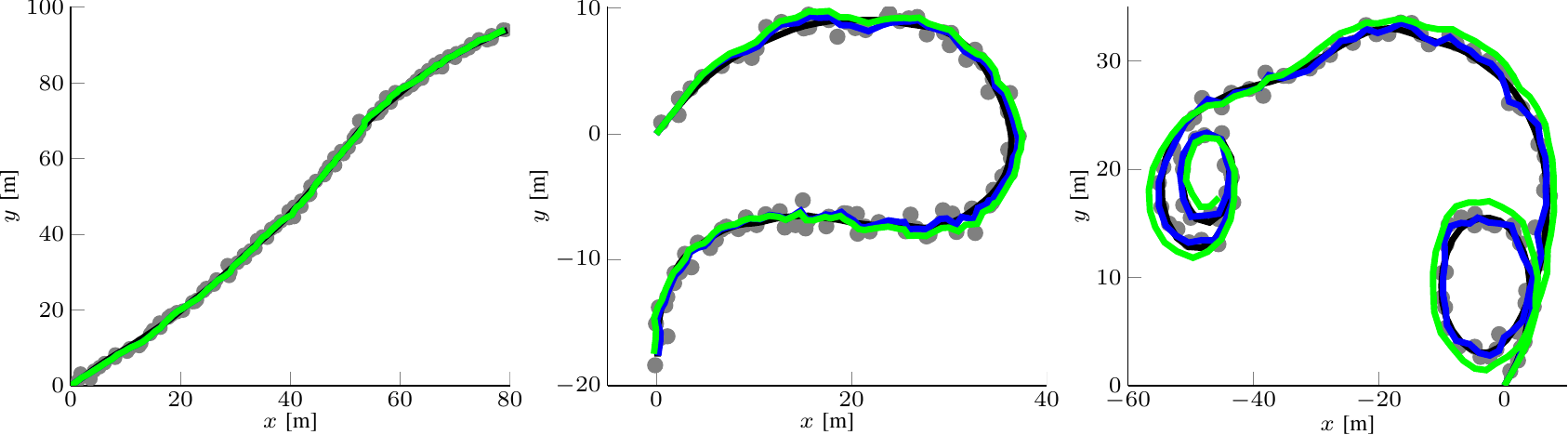}
\caption{
Examples of three different simulated trajectories, generated with the $\SE(2)^2$ motion model, with different intensities of process noise over rotational components, i.e., standard deviation in rotational component was $\sigma_{\omega} = [ 0.01 \ 0.1 \ 1 ]^{\circ}$. The blue line corresponds to $\SE(2)^2$ filter, while the green line represents the CV model ($\SE(2) \times \R^3$ and CTRV are omitted for clarity).}
\label{fig:comp_w}
\vspace{-0.0cm}
\end{figure*}
\noindent Let $\Omega_{1,k}$, $\Omega_{2,k}$ and $\Omega_{3,k}$ denote the first three rows of the vector \eqref{eq:motionG}, respectively (whereas the last three rows are trivial $\Omega_{4,k} = \Omega_{5,k} = \Omega_{6,k} = 0$). 
\begin{figure*}[!t]
\begin{equation}
\begin{split}
\dfrac{\partial \Omega_{1,k}}{\partial \epsilon_{v_x}} |_{\epsilon=0} &= 
\dT \cos \omega_k \frac{\sin \epsilon_{\omega}}{\epsilon_{\omega}} -
\dT \sin \omega_k \frac{\cos \epsilon_{\omega} - 1}{\epsilon_{\omega}} |_{\epsilon=0} = \dT \cos \omega_k
\\
\dfrac{\partial \Omega_{1,k}}{\partial \epsilon_{v_y}} |_{\epsilon=0} &= 
\dT \cos \omega_k \frac{\cos \epsilon_{\omega} - 1}{\epsilon_{\omega}} -
\dT \sin \omega_k \frac{\sin \epsilon_{\omega}}{\epsilon_{\omega}} |_{\epsilon=0} = -\dT \sin \omega_k
\\
\dfrac{\partial \Omega_{1,k}}{\partial \epsilon_{\omega}} |_{\epsilon=0} &=
\dT \cos \omega
\dfrac{
(\epsilon_{v_x} \cos \epsilon_{\omega} - \epsilon_{v_y} \sin \epsilon_{\omega}) \epsilon_{\omega} -
[\epsilon_{v_x} \sin \epsilon_{\omega} + \epsilon_{v_y} (\cos \epsilon_{\omega} - 1)]
}{\epsilon_{\omega}^2} \\
& - \dT \sin \omega
\dfrac{
(\epsilon_{v_x} \sin \epsilon_{\omega} + \epsilon_{v_y} \cos \epsilon_{\omega}) \epsilon_{\omega} -
[\epsilon_{v_x} (\cos \epsilon_{\omega} - 1) + \epsilon_{v_y} \sin \epsilon_{\omega}]
}{\epsilon_{\omega}^2} |_{\epsilon=0} = 0 \\
\dfrac{\partial \Omega_{2,k}}{\partial \epsilon_{v_x}} |_{\epsilon=0} &= \dT \sin \omega_k 
\,, \quad 
\dfrac{\partial \Omega_{2,k}}{\partial \epsilon_{v_y}} |_{\epsilon=0} = \dT \cos \omega_k
\,, \quad 
\dfrac{\partial \Omega_{2,k}}{\partial \epsilon_{\omega}} |_{\epsilon=0} = 0 \\
\dfrac{\partial \Omega_{3,k}}{\partial \epsilon_{v_x}} |_{\epsilon=0} &= 0
\,, \quad 
\dfrac{\partial \Omega_{3,k}}{\partial \epsilon_{v_y}} |_{\epsilon=0} = 0
\,, \quad 
\dfrac{\partial \Omega_{3,k}}{\partial \epsilon_{\omega}} |_{\epsilon=0} = \dT
\end{split}
\label{eq:derC}
\end{equation}
\vspace{-0.1cm}
\end{figure*}
Even though the multivariate limits
$\dfrac{\partial \Omega_{1,k}}{\partial \epsilon_{\omega}} |_{\epsilon=0}$ and 
$\dfrac{\partial \Omega_{2,k}}{\partial \epsilon_{\omega}} |_{\epsilon=0}$ 
appear involved, their derivation follow from patient algebraic manipulations.
The resulting terms are shown in \eqref{eq:derC}.
The matrix $\mathscr{C}_k$ is finally then given as
\begin{equation}
\label{eq:calCFinal}
\mathscr{C}_k =
\begin{bmatrix}
\0^{3 \times 3} &
\begin{matrix}
\dT\cos\omega_k & -\dT\sin\omega_k & 0 \\
\dT\sin\omega_k & \dT\cos\omega_k & 0 \\
0 & 0 & \dT
\end{matrix} \\
\0^{3 \times 3} & \0^{3 \times 3} 
\end{bmatrix}\,.
\end{equation}
The adjoint operators $\Ad_{\G}$ and $\ad_{\G}$ are formed block diagonally as 
\begin{equation}
\begin{split}
\Ad_{\G}(X) & = \text{diag} 
\left(
\Ad_{\SE(2)}(T_s), \,
\Ad_{\SE(2)}(T_d)
\right) \,, \\[1mm]
\ad_{\G}(\x) & = \text{diag} 
\left(
\ad_{\SE(2)}(\x_s), \,
\ad_{\SE(2)}(\x_d)
\right) \,.
\end{split}
\end{equation}

The last needed ingredient is the process noise covariance matrix $Q_k$.
Assuming the constant acceleration over the sampling period $\dT$, we model the process noise as a discrete white noise acceleration over the three components: $n_{x_k}$, $n_{y_k}$ and $n_{\omega_k}$.
At this point, we can use the equation \eqref{eq:pred_cov} for predicting the covariance of the system.

\begin{comment}
The careful reader may note that there exists a strong coupling between the choice of $\hat{\Omega}$ and type of product of the subspaces. As was already mentioned, in contrary to the case of an Euclidean space, there are several ways to combine Lie groups (e.g. direct product, semi-direct product, twisted product). For our particular problem, instead of choosing the element $\SE(2)$, we could have utilized the direct product $\SO(2) \times \R^2$. As an outcome, a different version of a motion model would show up
\begin{equation}
\begin{split}
X^*_{k+1} = X^*_{k} \, \exp_{\G^*} \left( [\Omega^*(X^*_k,n_k)]^{\wedge}_{\G^*} \right) \,,
\end{split}
\end{equation}
where $X^* \in  \G^* = \SO(2) \times \R^2 \times \SO(2) \times \R^2$ and $\Omega^*(X^*,n)$ is the corresponding nonlinear motion model.
Furthermore, in this case it is not possible to represent the model of the system with the additive noise in $\R^p$--space as assumed in \eqref{eq:model}. So the choice of a group product and corresponding adequate motion model is strongly related to both ease of implementation and gained performance of the filter. However, further discussion on this topic is out of the scope of this paper and the interested reader is referred to \cite{Chirikjian2012b}.
\end{comment}

\subsubsection{Update}
\label{par:update}
The predicted system state is described with $X_{k+1|k} \sim \mathcal{G}(\mu_{k+1|k},P_{k+1|k})$ and now we proceed to updating the state by incorporating the newly arrived measurement $Z_{k+1} \in \G'$.
In this case, we choose the measurements to arise in the Euclidean space $\R^2$, measuring the current position of the tracked object in $2$D.
This choice is application related and is more discussed in the next section.
For this reason and since the Euclidean space is a trivial example of a matrix Lie group, we introduce the
representation of $z = \begin{bmatrix} x_z \ y_z \ \end{bmatrix}^T \in \R^2$ in the form of a matrix Lie group $Z \in
\G' \subset \R^{3 \times 3}$ and Lie algebra $[z]^{\wedge}_{\R^2} \in \g' \subset \R^{3 \times 3}$
\begin{equation}
\label{eq:r}
Z = 
\begin{bmatrix}
\I^{2 \times 2} & z\\
\0^{1 \times 2} & 1
\end{bmatrix} \ \text{ and } \
[z]^{\wedge}_{\R^2} = 
\begin{bmatrix}
\0^{2 \times 2} & z\\
\0^{1 \times 2} & 0
\end{bmatrix} \,.
\end{equation}
Please note there exists a trivial mapping between the members of the triplet $\R^2$, $\g'$ and $\G'$, hence the formal inverses of the terms from \eqref{eq:r} are omitted here. 

The measurement function is the map $h : \SE(2)^2 \rightarrow \R^2$. The element that specifically needs to be derived
is the measurement matrix $\mathcal{H}_{k+1}$, which in the vein of \eqref{eq:derC}, requires using partial derivatives
and multivariate limits. 
Again, we start with definition of the Lie algebraic error 
$ \epsilon =
\begin{bmatrix}
\epsilon_x \ \epsilon_y \ \epsilon_{\theta} \ \epsilon_{v_x} \ \epsilon_{v_y} \ \epsilon_{\omega}
\end{bmatrix} $.
The function to be partially derived is given as
\begin{equation}
\begin{split}
&\hspace*{-0.2cm} \left[
    \log_{\G'} 
    \left(
    h(\mu_{k+1|k})^{-1} h\left( \mu_{k+1|k} \exp_{\G}  \left( [\epsilon]^{\wedge}_{\G} \right)
    \right)
    \right)
    \right]^{\vee}_{\G} = \\
&
\hspace*{1cm} 
\begin{bmatrix}
\cos\theta_{k+1|k} \, p_1 - 
\sin\theta_{k+1|k} \, p_2 \\
\sin\theta_{k+1|k} \, p_1 +
\cos\theta_{k+1|k} \, p_2
\end{bmatrix}  \,,
\end{split}
\label{eq:H_err}
\end{equation}
where
\begin{equation}
\begin{split}
p_1&=
\left[\epsilon_{x} \sin \epsilon_{\theta} + \epsilon_{y} (\cos \epsilon_{\theta} - 1)\right]\epsilon_{\theta}^{-1} \\
p_1&=
\left[\epsilon_{x} (1 - \cos \epsilon_{\theta}) + \epsilon_{y} \sin \epsilon_{\theta}\right]\epsilon_{\theta}^{-1}\,.
\end{split}
\end{equation}
Let $\H_{1,k+1}$ and $\H_{2,k+1}$ denote the two rows of expression \eqref{eq:H_err}.
In order to derive \eqref{eq:calH}, we need to determine partial derivatives and multivariate limits over all directions
of the Lie algebraic error vector, and the result is given in \eqref{eq:derH}.
\begin{figure*}[!t]
\begin{equation}
\begin{split}
\dfrac{\partial \H_{1,k+1}}{\partial \epsilon_{x}} |_{\epsilon=0} &= 
\cos \theta_{k+1|k} \frac{\sin \epsilon_{\theta}}{\epsilon_{\theta}} -
\sin \theta_{k+1|k} \frac{\cos \epsilon_{\theta} - 1}{\epsilon_{\theta}} |_{\epsilon=0} = \cos \theta_{k+1|k}
\\
\dfrac{\partial \H_{1,k+1}}{\partial \epsilon_{y}} |_{\epsilon=0} &= 
\cos \theta_{k+1|k} \frac{\cos \epsilon_{\theta} - 1}{\epsilon_{\theta}} -
\sin \theta_{k+1|k} \frac{\sin \epsilon_{\theta}}{\epsilon_{\theta}} |_{\epsilon=0} = - \sin \theta_{k+1|k}
\\
\dfrac{\partial \H_{1,k+1}}{\partial \epsilon_{\theta}} |_{\epsilon=0} &= 
\cos \theta_{k+1|k}
\dfrac{
(\epsilon_{x} \cos \epsilon_{\theta} - \epsilon_{y} \sin \epsilon_{\theta}) \epsilon_{\theta} -
[\epsilon_{x} \sin \epsilon_{\theta} + \epsilon_{y} (\cos \epsilon_{\theta} - 1)]
}{\epsilon_{\theta}^2} \\
&- \sin \theta_{k+1|k}
\dfrac{
(\epsilon_{x} \sin \epsilon_{\theta} + \epsilon_{y} \cos \epsilon_{\theta}) \epsilon_{\theta} -
[\epsilon_{x} (\cos \epsilon_{\theta} - 1) + \epsilon_{y} \sin \epsilon_{\theta}]
}{\epsilon_{\theta}^2} |_{\epsilon=0} = 0  \\
\dfrac{\partial \H_{2,k+1}}{\partial \epsilon_{x}} |_{\epsilon=0} &= \sin \theta_{k+1|k}
\,, \quad 
\dfrac{\partial \H_{2,k+1}}{\partial \epsilon_{y}} |_{\epsilon=0} = \cos \theta_{k+1|k}
\,, \quad 
\dfrac{\partial \H_{2,k+1}}{\partial \epsilon_{\theta}} |_{\epsilon=0} = 0\\
\end{split}
\label{eq:derH}
\end{equation}
\vspace{-0.1cm}
\end{figure*}
The final measurement matrix $\mathcal{H}_{k+1}$ amounts to
\begin{equation}
\label{eq:calH_model}
\mathcal{H}_{k+1} =
\begin{bmatrix}
\cos\theta_{k+1|k} & -\sin\theta_{k+1|k} & 0 & 0 & 0 & 0\\
\sin\theta_{k+1|k} & \cos\theta_{k+1|k} & 0 & 0 & 0 & 0
\end{bmatrix} \,.
\end{equation}
Again, the interested reader is directed to perform algebraic manipulations when calculating the multivariate limits for
proving \eqref{eq:calH_model}.
Here we deal with rather simple and most common measurement space, but as well as in some recent works \cite{Chirikjian2014}, the filter from Section~\ref{sec:filter} enables us to incorporate nonlinear measurements if needed.

Now we have all the means for updating the filter by calculating the Kalman gain $K_{k+1}$ and the innovation vector $\nu_{k+1}$ \eqref{eq:K}, and finally correcting the mean $\mu_{k+1}$ \eqref{eq:up_mu} and the covariance matrix $P_{k+1}$ \eqref{eq:up_P}. 

\subsection{Simulation}

In order to test the performance of the proposed filters, we have simulated trajectories of a maneuvering object in 2D,
where the motion of the system was described by the $\SE(2)\times \R^3$ and $\SE(2)^2$ models.
%We chose this model since it is the most general one among the tested models, capable of simulating a variety of types
%of motion.
Three examples of generated trajectories with the $\SE(2)^2$ model, with different levels of rotational process noise, are given in
Fig.~\ref{fig:comp_w}.
In order to test performance of the proposed filters, we conducted statistical comparison of $\SE(2)\times \R^3$ and $\SE(2)^2$, with two conventional approaches, i.e., (i) the EKF based constant turn rate and velocity
and (ii) the KF based CV models.

\begin{figure}[!t]
\centering
\vspace{0.2cm}
%  \tikzsetnextfilename{stats_SE2R3}
%  \input{figures/stats_SE2R3.tikz.tex}
\includegraphics{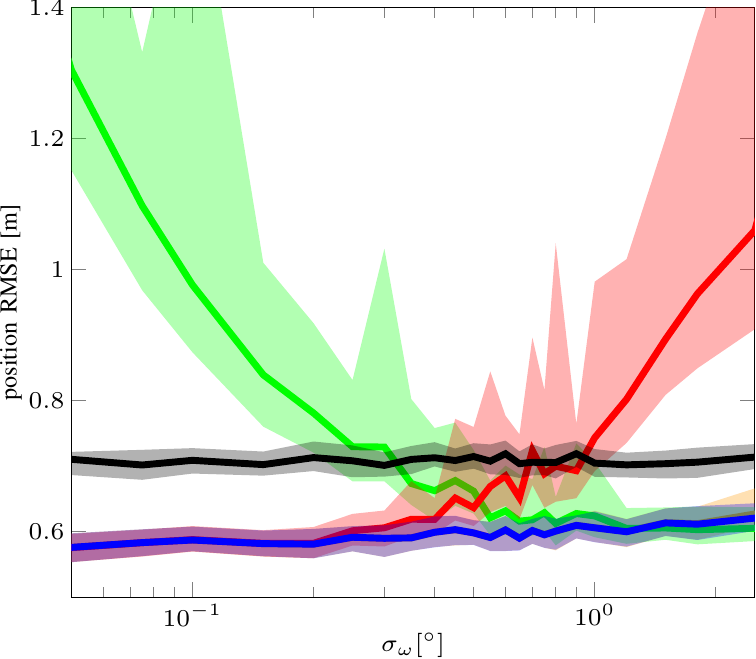}
\caption{\vspace{-0.0cm}
Performance statistics obtained over $100$ generated trajectories for $30$ different values of $\sigma_{\omega}$.
We have compared the proposed filter $\SE(2)^2$ (blue) and $\SE(2) \times \R^3$ (orange) with the EKF based CTRV (green), KF based CV (red), and measurements (black), where the solid lines
corresponds to mean values, while transparent areas correspond to
one standard deviation (in both $+/-$ directions) of each of the associated RMSEs. We can notice that the $\SE(2) \times \R^3$ and $\SE(2)^2$ filters, whose difference is barely noticable, exhibit similar behaviour, outperforming the other two filters.
}
\label{fig:statistics_plot}
\vspace{-0.2cm}
\end{figure}

The noise parameters that generated the trajectories were set as follows: $n_{v_x} \sim \mathcal{N}(0,0.1^2)$, $n_{v_y} \sim
\mathcal{N}(0,0.1^2)$, $n_{\omega} \sim \mathcal{N}(0,\sigma_{\omega}^2) $, where $\sigma_{\omega}$ took
$30$ equidistant values in the interval $[0,3]$. 
For each of these values of $\sigma_{\omega}$ we have generated $100$ trajectories and compared
the performance of the four filters.
The measurement noise was set to $m_x \sim \mathcal{N}(0,0.5^2)$ and $m_y \sim \mathcal{N}(0,0.5^2)$.
Special attention was given to parametrization of process noise covariance matrices in order to make the comparison as
fair as possible.
%\im{Napisati ovdje koliko je bio mjerni šum i komentirati kako je paženo da svi šumovi u filtrima budu jednaki koliko je to bilo moguće.}
Statistical evaluation of the root-mean-square-error (RMSE) in object's position is depicted in
Fig.~\ref{fig:statistics_plot}.
It can be seen that the $\SE(2)^2$ and $\SE(2)\times\R^3$ filters significantly outperform the other filters.
Specifically, when the rotation is not very dynamic, the KF based CV filter follows the trajectories well, while
with the increase in $\sigma_{\omega}$ its performance drops significantly.
On the contrary, when the rotation is not very dynamic, the EKF based CTRV filter struggles to follow the
trajectories correctly, while with the increase in $\sigma_{\omega}$ its performance gets closer to the one of
the proposed filters.

Considering the varying dynamism in the rotation, we assert that the $\SE(2)\times \R^3$ and $\SE(2)^2$ show very similar behaviour, while significantly outperforming the other two filters.
Particularly, they present the best of the two worlds: the CV and the CTRV behaviour.
Here we present statistical evaluation conducted on the trajectories generated by the $\SE(2)^2$ model,
Results on the trajectories generated by the $\SE(2)\times \R^3$ model showed similar inter-performance,
they are omitted from the present paper.
%However, we believe that the $\SE(2)^2$ model can find its application in the cases when the system kinematics allow
%\emph{banana-shaped} uncertainties both in the position
%and the velocity components, which, as was mentioned in Section~\ref{sec:mot}, can truly be exhibited by certain
%omnidirectional mobile platforms.
Furthermore, in simulations we only measured the position, i.e., the measurement space was in $\R^2$, while measuring
additionally the orientation, i.e., making the measurement space $\SE(2)$, would only further highlight the potential of
the $\SE(2)^2$ filter.
Both of the presented omnidirectional motion models are proven to be very flexible and capable of capturing various types of
motion that can be encountered in, e.g., busy intersection consisting of cars, trams, bicycles, motorcycles, and
pedestrians or an unknown environment that a robot enters for the first time consisting of different robot platforms and
humans.

%\im{So the idea of this comparison is not to show that $\SE(2)^2$ generally achieves better performance than $\SE(2) \times
%\R^3$ but this depends on the behaviour of the tracked system itself.}

\section{Conclusion}
\label{sec:con}\noindent
In this paper we have proposed novel models for tracking a moving object exploiting its motion on the rigid
body motion group $\SE(2)$.
The proposed filtering approach relied on the extended Kalman filter for matrix Lie groups, since the rigid body motion
group itself is a matrix Lie group.
Therefore, we have modeled the state space as either a direct product of the of the $\SE(2)$ group and the $\R^3$
vector, i.e., $\SE(2) \times \R^3$, or two $\SE(2)$ groups, i.e. $\SE(2) \times \SE(2)$, where the first term described
the current pose, while the second term handled second order dynamics.
We have analyzed the performance of the proposed filters on a large number of synthetic trajectories and compared them to
(i) the EKF based constant velocity and turn rate and (ii) the KF based constant velocity models.
The $\SE(2) \times \R^3$ and $\SE(2)^2$ filters showed similar performance on the synthetic dataset, and have significantly outperformed other well-established approaches for a wide range of intensities in the rotation component.
%the best performance for a wide range of intensities in the rotation.
%Due to the merits that the direct filtering on manifolds brings, the proposed filters has shown the best performance over a high range of trajectory dynamics.
%In order to tackle real-world tracking problems with clutter, we have also proposed the framework for employing probabilistic data association method on matrix Lie Groups and have applied it for the tracking problem.

Even though the presented work was applied on a tracking problem, we believe it can serve as a starting point for
further exploitation of estimation on matrix Lie groups and its applications on different problems.
The use of higher order dynamics may be of special interest for the domain of robotics, as well as for multi-target
tracking applications.
Furthermore, these techniques could also find application in other rigid body motion estimation problems requiring
precise pose estimation and higher-order motion.

\section*{Acknowledgments}
This work has been supported from the Unity Through Knowledge Fund under the project Cooperative Cloud based Simultaneous
    Localization and Mapping in Dynamic Environments (cloudSLAM) and the European Union’s Horizon 2020 research and innovation
    programme under grant agreement No 688117 (SafeLog).

%\newpage

\vspace{7mm}

\balance
\bibliographystyle{IEEEtran}
\bibliography{bibliography/EstimationManifold,bibliography/bibliography,bibliography/library}

\end{document}